\pdfoutput=1

\documentclass[11pt]{article}

\usepackage[preprint]{acl}
\usepackage{textcomp}

\usepackage[ruled]{algorithm2e}

\usepackage{times}
\usepackage{latexsym}
\usepackage{array}
\usepackage{multirow}
\usepackage{array}
\usepackage{CJKutf8}
\usepackage{longtable}
\usepackage{graphicx}
\usepackage{amsmath}

\usepackage[T1]{fontenc}

\usepackage[utf8]{inputenc}

\usepackage{microtype}
\usepackage{booktabs} 

\usepackage{inconsolata}
\usepackage{multirow}

\usepackage{graphicx}

\title{DeFine: A Decomposed and Fine-Grained Annotated Dataset \\ for Long-form Article Generation}

\author{Ming Wang\textsuperscript{1}\quad 
Fang Wang \textsuperscript{2}\quad 
Minghao Hu\textsuperscript{4}\textsuperscript{*}\quad 
Li He\textsuperscript{1}\textsuperscript{*}\\ 
\textbf{Haiyang Wang}\textsuperscript{3}\quad 
\textbf{Jun Zhang}\textsuperscript{4}\quad
\textbf{Tianwei Yan}\textsuperscript{3} \quad 
\textbf{Li Li}\textsuperscript{1}\quad
\textbf{Zhunchen Luo} \textsuperscript{4}\quad \\ 
\textbf{Wei Lu}\textsuperscript{4}\quad 
\textbf{Xiaoying Bai}\textsuperscript{4} \quad
\textbf{Guotong Geng}\textsuperscript{4} \\
\textsuperscript{1}North China University of Technology\quad   \textsuperscript{2} Peking University\\ \textsuperscript{3} National University of Defense Technology \quad \textsuperscript{4} Center of Information Research, AMS}

\begin{document}
\maketitle
\begin{abstract} Long-form article generation (LFAG) presents challenges such as maintaining logical consistency, comprehensive topic coverage, and narrative coherence across extended articles. Existing datasets often lack both the hierarchical structure and fine-grained annotation needed to effectively decompose tasks, resulting in shallow, disorganized article generation. To address these limitations, we introduce \textbf{DeFine}, a \textbf{De}composed and \textbf{Fine}-grained annotated dataset for long-form article generation. DeFine is characterized by its hierarchical decomposition strategy and the integration of domain-specific knowledge with multi-level annotations, ensuring granular control and enhanced depth in article generation. To construct the dataset, a multi-agent collaborative pipeline is proposed, which systematically segments the generation process into four parts: Data Miner, Cite Retreiver, Q\&A Annotator and Data Cleaner. To validate the effectiveness of DeFine, we designed and tested three LFAG baselines: the web retrieval, the local retrieval, and the grounded reference. We fine-tuned the Qwen2-7b-Instruct model using the DeFine training dataset. The experimental results showed significant improvements in text quality, specifically in topic coverage, depth of information, and content fidelity. Our dataset~\footnote{https://github.com/DeFine-LFAG/DeFine\_Dataset} publicly available to facilitate future research.

\end{abstract}

\footnotetext{\textsuperscript{*}Minghao Hu and Li He are the corresponding authors. Contact: \texttt{humh573@163.com} (Minghao Hu), \texttt{heli@ncut.edu.cn} (Li He).}



\section{Introduction}
The task of Long-Form Article Generation (LFAG) aims to enable large language models to automatically generate articles that are comprehensive, coherent, and factually accurate on various topics ~\cite{kasneci2023chatgpt,bubeck2023sparks}. This task requires models not only to cover relevant topics in depth, but also to maintain consistency in logical structure and narrative flow ~\cite{chen2023beyond}. However, achieving high-quality long-form article generation remains a significant challenge due to the complexity of handling large volumes of information and maintaining coherence over extended texts.

\begin{figure}[!t]
  \includegraphics[width=\columnwidth]{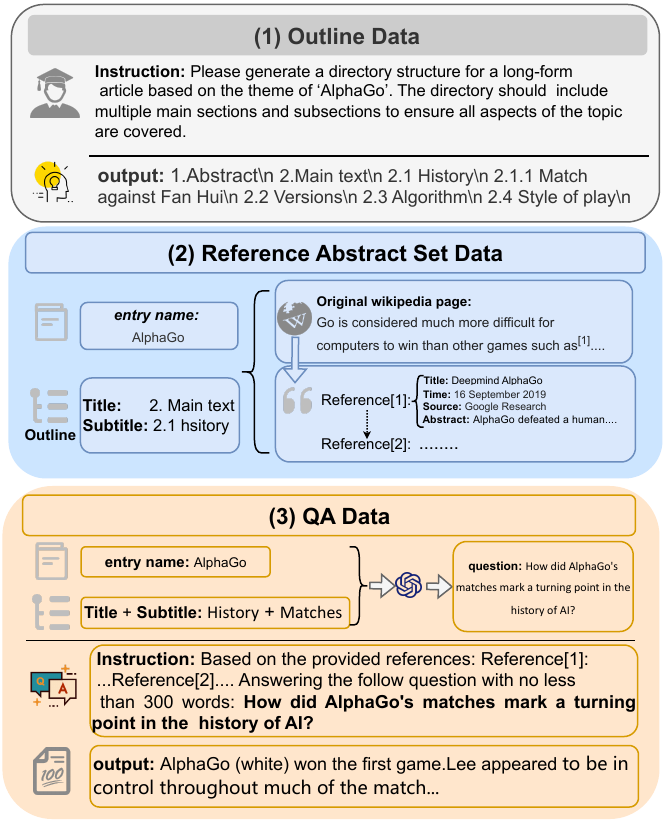}
  \caption{DeFine dataset example. In our dataset, there are three distinct types of data: outline data, abstract set data, and question-answer data. These are utilized to train the model for different functions: generating outlines based on a given article topic, extracting summaries from retrieved references, and generating long-form text based on the extracted summaries and outlines.}
  \label{fig:dataset examples}
\end{figure}

With the advent of large-scale pre-trained language models such as Gemini ~\citep{gemini2023} and GPT-4 ~\citep{gpt4paper}, significant advances have been made in long-form article generation. For instance, STORM ~\citep{shao2024assisting} improves the quality of long-form article generation through a two-stage generation strategy; Co-STORM ~\citep{jiang-etal-2024-unknown} introduces a collaborative dialogue mechanism, employing a round-based management strategy to facilitate AI agent discussions and collaborative research; AGENTS' ROOM ~\citep{huot2024agentsroomnarrativegeneration} utilizes intelligent agent technology to simulate interactions among experts for the generation of long-form articles. Despite these advancements, these methods rely heavily on prompt engineering and AI agents, lacking datasets specifically designed for LFAG tasks to enhance fine-tuning.

A significant bottleneck in LFAG research is the absence of high-quality datasets specifically designed for both outline construction and detailed long-form article generation. Existing datasets like LongLaMP \citep{kumar2024longlampbenchmarkpersonalizedlongform}, while providing a benchmark for personalized long-text generation, focus primarily on individualization rather than structured article synthesis. Similarly, LongWriter \citep{bai2024longwriterunleashing10000word} enables ultra-long text generation but lacks explicit hierarchical structure and task decomposition crucial for long-form article construction. In addition, HelloBench \citep{que2024hellobenchevaluatinglongtext} provides an evaluation benchmark designed to assess LLMs' ability to generate coherent long-form texts across multiple genres and tasks, yet it serves only as an evaluation tool without providing training data for model improvement. Finally, while ELI5 \citep{fan2019eli5longformquestion} and ASQA \citep{stelmakh2023asqafactoidquestionsmeet} encourage long, detailed, and well-grounded responses for question-answer tasks, they do not offer the structured, multi-step annotations necessary for fine-grained control over long-form article generation.



In this work, we address these challenges by introducting DeFine, a hierarchically decomposed and fine-grained annotated dataset specifically designed to support long-form article generation. DeFine includes: (1) a structured breakdown of the generation process into distinct three stages: outline creation, reference retrieval and extraction, question-answer data generation, ensuring logical coherence and content organization, and (2) fine-grained annotations at each stage, ensuring logical coherence and content organization. As shown in Figure \ref{fig:dataset examples}, DeFine provides detailed supervision and control over generated content, enabling better management of task decomposition and model performance.



To construct DeFine efficiently, we employ a multi-agent collaboration pipeline, where each agent specializes in a specific aspect of dataset creation. This pipeline is a combination of: (1) Data Miner: extracts and organizes structural titles and subtitles from high-quality articles to construct hierarchical outline data. (2) Cite Retriever: retrieves referenced URLs and refines the data into content-abstraction sets. (3) Q\&A Annotator: generates question-answer data by applying hallucination detection algorithms and context-aware prompts. (4) Data Cleaner: ensures data integrity through rigorous cleaning from multiple perspectives, including richness, relevance, and coverage.  


Our main contributions are as follows:


1) We introduce DeFine, a novel hierarchical decomposed and fine-grained annotated dataset designed to enhance long-form article generation;

2) We propose a multi-agent collaboration pipeline to construct long-form article generation dataset.

3) We validate the effectiveness of DeFine through the application of three LFAG baselines: web retrieval, local retrieval, and grounded reference. Experimental results show that models fine-tuned with DeFine outperform existing LFAG approaches in logical coherence, factual accuracy, and citation reliability.

\section{Related Work}
\subsection{LFAG Datasets}
High-quality datasets are essential for advancing research on Long-Form Article Generation (LFAG). Within the question-answering (QA) domain, several existing datasets focus on long-form answers, aiming to generate detailed and well-grounded responses ~\cite{fan2019eli5longformquestion,stelmakh2023asqafactoidquestionsmeet,cohen2021wikisum,jin2019pubmedqa,kovcisky2018narrativeqa}. While these datasets encourage lengthier outputs, they lack structural decomposition and hierarchical content organization, making them insufficient for complex article generation. In addition, LongLaMP \citep{kumar2024longlampbenchmarkpersonalizedlongform} provides a benchmark for personalized long-text generation, capturing user-specific writing styles but without a structured decomposition process. LongWriter \citep{bai2024longwriterunleashing10000word} extends LLM output length but does not provide task-level decompositions. LongCite \citep{zhang2024longciteenablingllmsgenerate} introduces sentence-level citation annotations, but focusing primarily on retrieval-based QA rather than structured article synthesis. 

To address these limitations, DeFine employs a clear hierarchical decomposition, enabling fine-grained control over article generation, with features like comprehensive outline creation, detailed reference use, and accurate citation metadata, ensuring coherent, factually consistent long-form content across various domains.


\subsection{Long-form Writing Methods}

Recent research has proposed various methods to tackle the complexities of long-form article generation (LFAG). Systems like STORM \cite{shao2024assisting} utilize large language models and structured references to enhance text quality through multi-stage generation. Similarly, CO-STORM \cite{jiang-etal-2024-unknown} introduces a collaborative dialogue mechanism, where AI agents engage in round-based discussions to co-create long-form content, enhancing coherence and depth. AGENTS' ROOM ~\cite{huot2024agentsroomnarrativegeneration} leverages intelligent agent to simulate expert-level interactions, improving the generation process by orchestrating discussions between domain experts. However, while these systems improve content quality, they still face difficulties in integrating diverse sources and maintaining logical coherence across extended pieces of text. AgentWrite \cite{bai2024longwriter} have optimized model architectures to extend the ability of LLMs to generate ultra-long articles, but challenges still persist in seamlessly integrating information from different sources and maintaining overall content coherence.


\subsection{Multi-agent Systems}
Recent advancements have highlighted the ability of LLM-based agents to exhibit strong reasoning and planning skills across a variety of domains \citep{zhao2024surveylargelanguagemodels,bubeck2023sparksartificialgeneralintelligence}. In multi-agent systems, multiple independent language models collaborate to solve complex tasks, often achieving results that surpass the capabilities of any single agent \citep{han2024llmmultiagentsystemschallenges,guo2024largelanguagemodelbased,talebirad2023multiagentcollaborationharnessingpower}. CO-STORM \cite{jiang-etal-2024-unknown} employs a multi-agent collaborative dialogue mechanism, where multiple AI agents work together in round-based interactions to jointly create long-form content. Similarly, AGENTS' ROOM \cite{huot2024agentsroomnarrativegeneration} utilizes intelligent agents to simulate expert-level interactions, enabling the system to conduct collaborative discussions between domain-specific agents. Chain-of-Agents (CoA) framework ~\cite{zhang2024chainagentslargelanguage} takes a similar multi-agent approach, where independent agents collaborate on long-context tasks to improve the coherence and relevance of generated text. Our dataset construction method align with these systems by employing a cooperative framework of specialized agents.




\section{Dataset DeFine} 
\begin{figure*}[t]
  \centering
  \includegraphics[width=\textwidth]{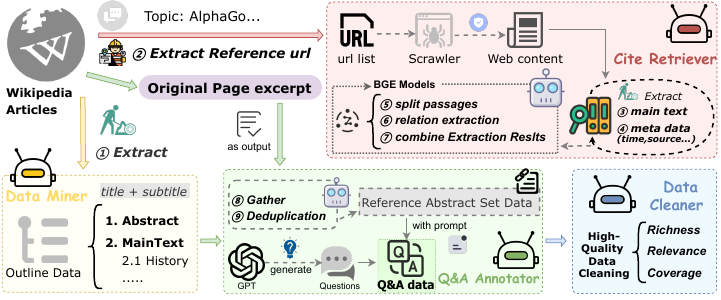}
  \caption{Overview of the DeFine dataset construction. We utilize four specialized agents to construct the dataset. First, the Data Miner extracts hierarchical outline data from articles. Next, the Cite Retriever summarizes reference content into abstract sets. The Q\&A Annotator generates question-answer pairs and applies hallucination detection. Finally, the Data Cleaner ensures the dataset’s quality through rigorous cleaning, focusing on richness, relevance, and coverage.}
  \label{fig:dataset_construction}
\end{figure*}

\subsection{Data Source} 
The DeFine leverages wikipedia\footnote{https://www.wikipedia.org/} as the primary data source due to its extensive article length, rich structured information (including hierarchical headings and paragraphs), and detailed citation references, which collectively facilitate the construction of the Long-Form Article Generation (LFAG) dataset. 


\subsection{Data Annotation}
Existing datasets suffer from a lack of hierarchical structure and task decomposition, and most only provide evaluation benchmarks or focus on personalized or question-answering tasks, lacking the fine-grained, multi-step annotations required for long-form text generation. To address the above issues, we built a multi-agent pipeline, which is broken down into the following manageable agents: \textbf{Data Miner}, \textbf{Cite Retriever}, \textbf{Q\&A Annotator} and \textbf{Data Cleaner}.

\subsubsection{Data Miner: Outline Data Extraction} 
This step leverages human-built web crawling techniques implemented by the Data Miner agent to meticulously extract structural elements—such as titles, subtitles, and hierarchical sections—from high-quality Wikipedia articles. By constructing a well-organized and detailed outline, Data Miner ensures that the extracted data maintains a clear and logical structure. 


\subsubsection{Cite Retriever: References \& Summary Retrieval}

The Cite Retriever agent leverages the extraction capabilities of the BGE-m3 relational extraction model ~\cite{chen2024bge} to parse and segment the content of Wikipedia articles, as well as intelligently retrieve and extract text information from cited URLs. Cite Retriever transforms raw data into refined content-abstract pairs that contain key information and contextual relationships.


\subsubsection{Q\&A Annotator: Question-Answer Data Generation} 
In this step, the Q\&A Annotator agent first performs precise sentence segmentation on both the abstract data and original text paragraphs, ensuring data operability and accuracy. Building on this foundation, Q\&A Annotator employs a sophisticated hallucination detection algorithm, utilizing entity coverage as a key filtering criterion to rigorously evaluate and enhance the quality and relevance of the abstract data. Once the data is refined, Q\&A Annotator dynamically designs multiple prompts for each paragraph, tailored to the length and context of the original text. These prompts, guided by carefully crafted templates (detailed in Appendix \ref{sec:appendixA}), enable the generation of high-quality question-answer pairs that are both contextually accurate and information-rich. 

\begin{algorithm}[b]
    \textit{\textbf{Generation}: Given the topic AlphaGo, and the subtitle History - Match against Fan Hui, please generate a question based on these two titles.}\\
    \textit{\textbf{Question}: How did the match between AlphaGo and Fan Hui mark a turning point in its history?}\\
    \textit{\textbf{Answer}: In its history...}\\
\end{algorithm}


The \textit{Question} in the prompts as follow are generated by the Qwen2-72b-Instruct model ~\citep{qwen}, which constructs questions around the themes of the original text and the corresponding subtitles of the paragraphs. For example, for the theme and subtitle \textit{"AlphaGo - History - Match against Fan Hui," }the model can generate questions like \textit{"How did the match between AlphaGo and Fan Hui mark a turning point in its history?"} Diversified prompts improve the diversity of the generated content, avoiding monotonous or repetitive outputs and covering more information points. Furthermore, they enhance the model's robustness by testing its performance in different contexts and identifying weaknesses for improvement.

\subsubsection{Data Cleaner: High-Quality Data Cleaning}
To ensure the highest quality of the dataset, the Data Cleaner agent conducts rigorous data cleaning from three distinct perspectives, filtering the data to guarantee its accuracy and consistency.

\begin{itemize}
    \item \textbf{Richness.} Data Cleaner establishes strict selection criteria, including minimum word count and a required number of references, to ensure that the selected Wikipedia entries are both detailed and information-rich..

    \item \textbf{Relevance.} Leveraging advanced techniques, Data Cleaner analyzes the references for each article to ensure they encapsulate the key information. Using a BGE-M3 model-based relation extraction method ~\citep{chen2024bge}, it identifies key entities and their relationships with the original text, enabling the extraction of relevant sentences and the formation of concise summaries. This process significantly improves precision while filtering out irrelevant or redundant content.

    \item \textbf{Coverage.} To further validate data integrity, we developed the Hallucination Detection Algorithm for Citation Reliability (HDACR). HDACR identifies entities in both reference and generated texts, evaluates matches, and detects hallucinations \citep{chen2024unifiedhallucinationdetectionmultimodal,zhang2024transferableefficientnonfactualcontent,li2024dawndarkempiricalstudy}. It generates a report with detection results, reference information, and unverifiable entities. By applying HDACR, we filter out references that inadequately cover the original text, enhancing dataset accuracy and text quality while providing detailed validation for long-form article generation. The detailed information about the HDACR algorithm and its pseudocode can be found in the appendix \ref{sec:appendix B}.

\end{itemize}

\subsection{Dataset Analysis}
\begin{figure}[!t]
  \includegraphics[width=\columnwidth]{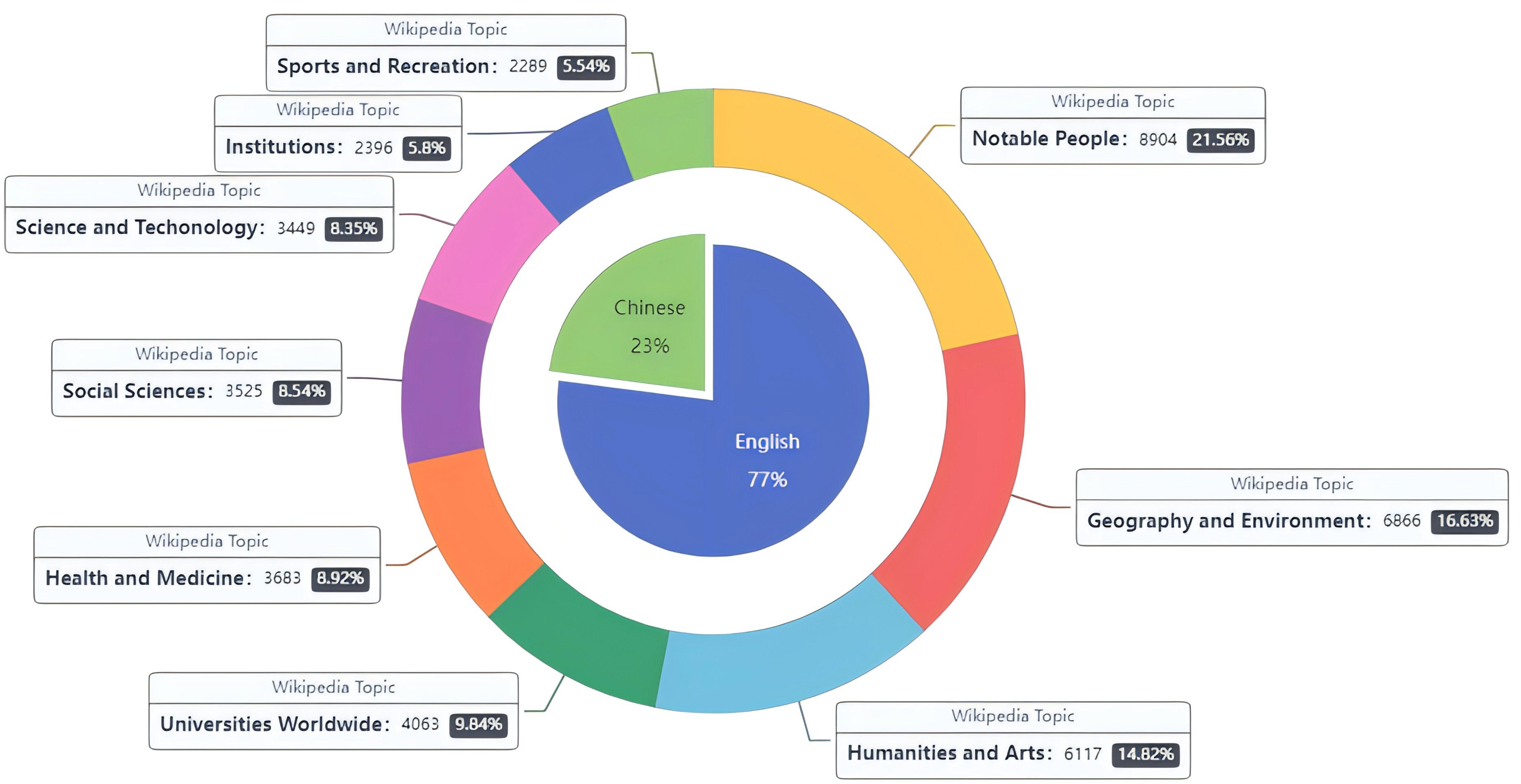} 

  \caption{Wikipedia pages distribution across different types.} 
  \label{fig:experiments}
\end{figure}

\textbf{Data Quantity and Categories.} DeFine contains 6,502 question-answer data, 9,647 abstract set data, and 52,045 outline data, covering both Chinese and English languages across various domains, including science, humanities, history, geography, literature, medicine, and sports. Figure \ref{fig:experiments} reflects the composition of the Wikipedia articles used to construct the dataset. English articles make up 77\% of the source material, while Chinese articles account for 23\%. In terms of subject matter sourced from Wikipedia, biographies have the highest proportion (21.56\%), followed by geography and environment (16.63\%), and humanities and arts (14.82\%). In contrast, sports and leisure articles, which also contribute to the dataset, have the lowest proportion, comprising only 5.54\% of the articles used.

\noindent\textbf{Answer Type and Length.} The outline data contains an average of 15 subheadings per entry, providing a detailed framework for generating structured long-form articles. For QA data, each entry includes an average of four reference summaries, with the maximum number of reference summaries reaching 42. The answer sections in the QA data exhibit an average word count of 375 words per answer. The longest answer contains 2,919 words, while the shortest consists of 150 words. This wide range in reference and answer lengths highlights the dataset’s diversity and richness, enabling models to generate both concise and elaborate long-form articles.

\noindent\textbf{Additional Comparison to LongWriter and LongLaMP}. LongWriter \citep{bai2024longwriter} focuses on generating ultra-long texts, extending the length significantly but without structured decomposition. LongLaMP \citep{kumar2024longlampbenchmarkpersonalizedlongform}, similarly, allows for long-form article generation but emphasizes personalization by tailoring outputs based on user-specific styles. However, it lacks the fine-grained hierarchical annotation necessary for coherent and well-structured long-form articles. In contrast to both LongWriter and LongLaMP, which may lack the structure necessary for cohesive long-form article generation, DeFine’s hierarchical decomposition provides a more controlled and accurate method for generating high-quality articles.

\section{Experimental Setup}
\subsection{LFAG baselines} 
We propose three LLM-based baselines, each optimized for different retrieval and generation strategies within a unified framework.

\textbf{\textit{Web Retrieval}}: This baseline integrates real-time web retrieval with a generation model to obtain the latest information. We utilize the Google SERPER API\footnote{https://google.serper.dev/search} as the retrieval engine and the Qwen2-7b-Scribe model, trained on the DeFine dataset, for generation. Both components are modular and can be replaced if necessary, ensuring high timeliness and dynamic data integration into generated articles.

\begin{figure*}[!htbp]
  \centering
  \includegraphics[width=\textwidth]{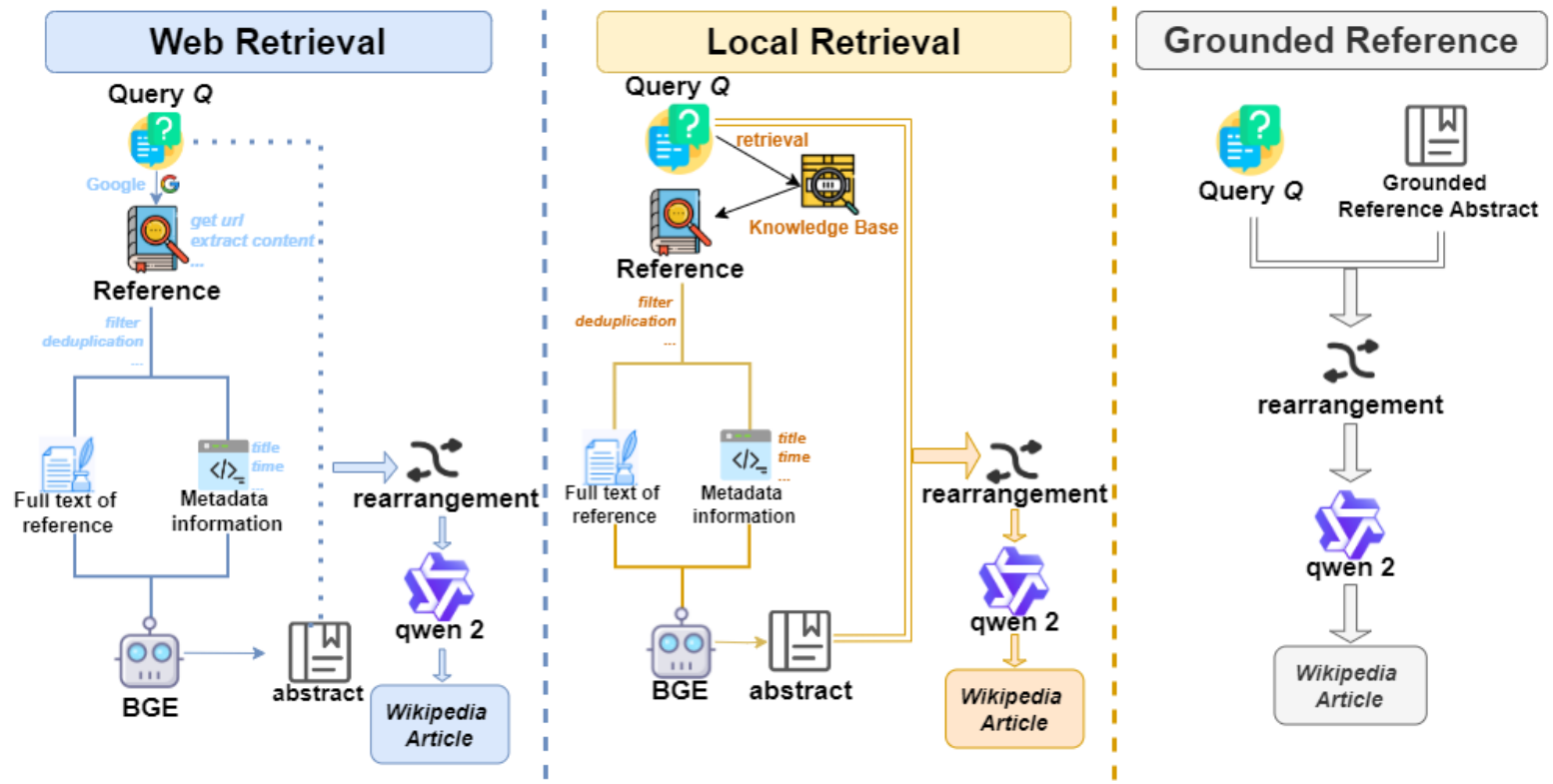}  
  \caption{Overview of the three baselines. Each baseline performs distinct tasks: the \textit{Web Retrieval} baseline integrates real-time web retrieval with a generation model to provide up-to-date information; the \textit{Local Retrieval} baseline retrieves references from a pre-built knowledge base and generates articles based on these references; the \textit{Grounded Reference} baseline directly uses original article abstracts to generate content. Notably, the relationship extraction model in all baselines uses the trained \textit{BGE-M3 model}, and both the generation model and the relationship extraction model are interchangeable, allowing for flexible adaptation to different research needs.}
  \label{fig:dataset_overview}
\end{figure*}

\textbf{\textit{Local Retrieval}}: This baseline retrieves references from a high-quality internal knowledge base built from Wikipedia articles. It identifies relevant content based on the article's topic and subheadings, extracts key information, and uses the model to generate detailed articles.

\textbf{\textit{Grounded Reference}}: This baseline directly inputs the article's topic, subheadings, and original Wikipedia abstracts into the generation model, without retrieval, to generate the complete content.

\subsection{Evaluation Metrics}

\textbf{\textit{Outline Evaluation.}} We employed the following two metrics to assess the quality of the outlines:

\begin{itemize}
    \item \textbf{Heading Soft Recall}. This metric measures the alignment between the generated outline and the reference outline at the thematic level, allowing for some degree of fuzzy matching.
    \item \textbf{Heading Entity Recall}. This metric evaluates the coverage of key entities (such as people and locations) in the generated outline, ensuring its overall quality.
\end{itemize}

\textbf{\textit{Long Articles Quality Evaluation.}} We employ two evaluations based on the availability of labels:
\begin{itemize}
    \item \textbf{Comparison with Human-Written Articles}. Using real long-form articles as a reference, we use \textit{ROUGE scores}~\citep{lin2004rouge} and calculate entity recall based on the paragraph-level FLAIR NER results.
    \item \textbf{Rubric Grading}. In the absence of real labels for the generated long articles, we use Prometheus~\citep{kim2023prometheus}, a 13B parameter evaluation model, which follows a rubric based on the STORM system~\citep{shao2024assisting}. It evaluates the quality of the article based on \textit{Interest Level}, \textit{Coherence and Organization}, \textit{Relevance and Focus}, and \textit{Coverage}.
\end{itemize}


\section{Results and Analysis}
\subsection{Main Results}

\begin{table*}[!h]
\centering
\resizebox{\textwidth}{!}{%
\begin{tabular}{c!{\vrule width \lightrulewidth}cc!{\vrule width \lightrulewidth}cc}
\toprule
\multirow{2}{*}{\textbf{Models}} & \multicolumn{2}{c|}{\textbf{DeFine}}        & \multicolumn{2}{c}{\textbf{FreshWiki}}      \\ \cmidrule(lr){2-3} \cmidrule(lr){4-5}
                                 & Heading Soft Recall & Heading Entity Recall & Heading Soft Recall & Heading Entity Recall \\ \midrule
Qwen2-7b-Instruct                & 65.15               & 47.18                 & 75.49                   & 34.28                     \\
Qwen1.5-7b-chat                  & 60.53               & 46.19                 & 73.32                   & 30.49                     \\
GPT 3.5                          & 77.62               & 36.66                 & 81.33               & 33.62                 \\
STORM                            & 78.13                   & 51.43                     & 86.26      & 40.52                 \\
Llama3-8b                        & 75.15               & 47.84                 & 74.31               & 32.15                 \\ \midrule
\textbf{Qwen2-7b-Scribe}         & \textbf{84.01}      & \textbf{71.16}        & 85.90     & \textbf{43.34}        \\ \bottomrule
\end{tabular}
}
\caption{Comparison of different models on two datasets, DeFine and FreshWiki. The Heading Soft Recall and Heading Entity Recall metrics are reported as percentages (\%). Values marked with bold indicate the best results. The \textit{STORM} results are taken from ~\citep{shao2024assisting}, while \textit{STORM} the results in my table are based on GPT-3.5.}
\label{tab:my-table1}
\end{table*}

\begin{table*}[!htbp]
\centering
\resizebox{\textwidth}{!}{%
\begin{tabular}{c!{\vrule width \lightrulewidth}ccc!{\vrule width \lightrulewidth}cccc}
\toprule
\multirow{2}{*}{\textbf{Models}} & \multicolumn{3}{c|}{\textbf{Comparison with Human-Written Articles}} & \multicolumn{4}{c}{\textbf{Rubric Grading}}\\ \cmidrule(lr){2-4} \cmidrule(lr){5-8}
                                 & ROUGE-1               & ROUGE-L               & Entity Recall           & Interest Level & Organization  & Relevance     & Coverage \\ \midrule
Direct Gen                       & 24.13                 & 11.08                 & 5.02                 & 2.35           & 4.23          & 3.02          & 3.98  \\ 
RAG                              & 27.31                 & 12.19                 & 7.82                 & 2.15           & 3.23          & 3.25          & 4.18  \\ 
oRAG                             & 43.26                 & 17.21                 & 14.27                & 3.83           & 4.81          & 4.19          & 4.35  \\
STORM                            & 44.79                 & 18.15                 & 13.03                & 3.87           & 4.81          & 4.43          & 4.62  \\ \midrule
GPT 3.5                          & 50.54                 & 36.24                 & 20.21                & 4.55           & 4.83          & 4.75          & 4.82  \\ \midrule
\textbf{Qwen2-7b-Scribe} & \textbf{53.39} & \textbf{37.54} & \textbf{22.85} & \textbf{4.85} & \textbf{4.92} & \textbf{4.82} & \textbf{4.87} \\ 
\bottomrule
\end{tabular}
}
\caption{Comparison of different models with human-written articles on several metrics. The ROUGE-1, ROUGE-L, and Entity Recall scores are reported as percentages (\%) to compare the quality of generated article, while the rubric grading evaluates Interest Level, Organization, Relevance, and Coverage, also expressed as percentages (\%). The results marked with bold indicate the best performance.}
\label{tab:my-table2}
\end{table*}

We fine-tuned the Qwen2-7b-Instruct model~\citep{yang2024qwen2technicalreport} using the DeFine training set. The fine-tuned Qwen2-7b-Instruct model, referred to as \textit{Qwen2-7b-Scribe}, was evaluated using 270 outline test samples and 270 QA test samples, and was further validated by generating full-length articles on DeFine test set and FreshWiki test set \citep{shao2024assisting}. Detailed information on DeFine dataset partitioning can be found in the appendix \ref{sec:appendixDatasetStats}.

\noindent\textbf{\textit{Outline Generation.}}
We evaluated the performance of the \textit{Qwen2-7b-Scribe, Qwen2-7b-Instruct} ~\citep{yang2024qwen2technicalreport}, \textit{Qwen1.5-7b-chat} ~\citep{qwen}, \textit{Llama-3-8b} ~\citep{llama3modelcard}, and \textit{gpt-3.5-turbo} models in generating outlines using the \textbf{DeFine} and \textbf{FreshWiki} datasets. The detailed results are presented in Table \ref{tab:my-table1}.

1) Qwen2-7b-Scribe demonstrates exceptional performance in the outline generation task. On two datasets, the model achieved an average improvement of 4.3\% to 14.4\% in heading soft recall and 5.9\% to 27.1\% in heading entity recall. These results indicate that the generated outlines are structurally well-aligned with the reference outlines. Furthermore, a further evaluation of the heading entity recall confirms the model's effectiveness in retaining key entities in the generated outlines. 

2) Due to differences in pretraining parameters and versions, the performance of Qwen1.5-7b-chat and Qwen2-7b-Instruct in outline generation is inferior to that of Qwen2-7b-Scribe. 

3) In comparison with the STORM model, the results show that Qwen2-7b-Scribe exhibits a higher heading entity recall, with its generated outlines retaining more key entities. However, Qwen2-7b-Scribe has a lower heading soft recall, which can be attributed to STORM's foundation on the GPT and its reliance on a more extensive pretraining corpus, granting it superior understanding capabilities.

\noindent\textbf{\textit{Long Article Generation. }}
We used the \textbf{DeFine} test set (270 article topics) and the \textbf{FreshWiki} test set~\citep{shao2024assisting} to generate articles via the web pipeline, and compared the results with the output of the STORM, as shown in Table \ref{tab:my-table2}.

1) Comparison with Human-Written Articles, which involves evaluation with real labels, shows that Qwen2-7b-Scribe achieves significant improvements in ROUGE-1 and entity recall. This indicates that the model effectively retains key entities during the generation of long article, successfully conveys and maintains important entity associations, and further enhances the quality and accuracy of the generated long article.

2) Rubric Grading, there is a lack of true labels. Qwen2-7b-Scribe shows significant improvement across all four evaluation metrics, demonstrating outstanding logical consistency and accuracy.

The effectiveness of the method is thoroughly validated by achieving high scores across all evaluation metrics, highlighting both the robustness of the dataset and the precision of the method.

\subsection{Ablation Study}

\begin{table*}[]
\centering
\resizebox{\textwidth}{!}{%
\begin{tabular}{cccccc}
\hline
Baseline & Model & ROUGE-1 & ROUGE-2 & ROUGE-L & Entity Recall \\ \hline
\multirow{4}{*}{Web Retrieval} & GPT-3.5 & 43.98 & 22.24 & 31.18 & 8.25 \\
& Qwen1.5-7b-chat & 53.13 & 30.57 & 39.15 & 9.36 \\
 & Qwen2-7b-Instruct & 50.46 & 31.50 & 39.78 & 8.96 \\
 & \textbf{Qwen2-7b-Scribe} & \textbf{53.93} & \textbf{32.41} & \textbf{39.80} & \textbf{14.09} \\ \hline
\multirow{4}{*}{Local Retrieval} & GPT-3.5 & 41.37 & 20.17 & 29.96 & 7.92 \\
 & Qwen1.5-7b-chat & 40.16 & 18.07 & 28.81 & 6.28 \\
 & Qwen-2-7b-Instruct & 42.52 & 21.44 & 30.42 & 8.54 \\
 & \textbf{Qwen-2-7b-Scribe} & \textbf{44.06} & \textbf{24.19} & \textbf{32.35} & \textbf{12.89} \\ \hline
\multirow{4}{*}{Grounded Reference} & GPT-3.5 & 44.41 & 22.41 & 31.26 & 8.90 \\
 & Qwen1.5-7b-chat & 46.42 & 23.11 & 32.15 & 10.35 \\
 & Qwen2-7b-Instruct & 46.70 & 23.74 & 32.47 & 9.38 \\
 & \textbf{Qwen2-7b-Scribe} & \textbf{50.79} & \textbf{26.82} & \textbf{34.93} & \textbf{13.61} \\ \hline
\end{tabular}
}
\caption{Comparison of different models on several ROUGE metrics (ROUGE-1, ROUGE-2, ROUGE-L) and entity recall. Results are shown for three baselines: web retrieval, local retrieval, and ground truth. The highest scores for each baseline are marked in bold, indicating the best performance achieved by the Qwen2-7b-Scribe model.}
\label{tab:my-table3}
\end{table*}

\begin{table}[ht]
\centering
\begin{tabular}{ccccc}
\toprule

 & Int. & Org. & Rel. & Cov. \\ \midrule
 Direct & 3.85 & 4.07 & 3.34 & 4.02 \\ 
 Local Retrieval & 4.83 & 4.78 & 4.71 & 4.55 \\ 
 \textbf{Web Retrieval} & \textbf{4.87} & \textbf{4.95} & \textbf{4.71} & \textbf{4.55} \\ \bottomrule
\end{tabular}
\caption{Results of Rubric Grading score for generated article quality evaluation. The Rubric Grading criteria (Int., Org., Rel., Cov.) are abbreviated for compactness.}
\label{tab:my-table4}
\end{table}

\textbf{\textit{The Impact of the Retrieval Process on Comparisons with Human-Written Articles. }}
In our study, the effectiveness of the retrieval process in long-form article generation is the primary focus, with experimental results presented in Table \ref{tab:my-table3}.

We introduced a "direct" pipeline in which all retrieval components are completely disabled. This setup allows the large language models (LLMs) to generate responses solely based on its pre-trained knowledge and the initial prompts, without any supplemental data retrieval. The aim is to test the model's core ability to generate coherent and relevant content without the aid of dynamic or static retrieval tools.

We evaluated \textit{Qwen2-7b-Scribe}, \textit{gpt-3.5-turbo}, and two other 7b models using a QA data test set, analyzing the generated article against human-written articles. Among the three test pipelines, \textit{Qwen2-7b-Scribe} demonstrated exceptional performance, particularly in the Web pipeline, where it leveraged extensive internet resources to outperform others in generating high-quality articles.

\noindent\textbf{\textit{The Impact of the Retrieval Process on Rubric Grading. }}
Our findings, presented in Table \ref{tab:my-table4}, indicate that models equipped with retrieval capabilities outperformed the "Direct" pipeline across all rubric criteria, confirming the hypothesis that dynamic data access significantly enhances content quality. This disparity highlighted the substantial role that retrieval plays in informing and structuring the generated content, especially in terms of relevance and depth.

This ablation study effectively demonstrates the critical importance of retrieval mechanisms in the LFAG tasks, providing clear evidence that the ability to dynamically integrate external information is pivotal for producing high-quality, accurate, and engaging long-form articles.

\section{Conclusion} 
We introduce DeFine dataset, a novel high-quality resource designed to enhance long-form article generation. By employing a hierarchical decomposition strategy and fine-grained annotation, we addressed key challenges in LFAG task, ensuring a structured approach to generating coherent and detailed articles. We evaluated our dataset across three LFAG baselines, and experimental results show that DeFine significantly improves logical coherence, information consistency, and citation accuracy in generated articles. Our work also demonstrates the effectiveness of integrating both retrieval-based and generation-based approaches, showing the potential of DeFine as a benchmark for future research in LFAG task.

\section*{Limitations}
Despite the significant contributions of the DeFine dataset to the field of long-form article generation, the current implementation presents certain limitations. One of the primary issues is the imbalance between the English and Chinese data, which stems from the limited number of Chinese Wikipedia entries compared to the far more extensive English Wikipedia. This may impact the model's ability to generalize across both languages, leading to potential performance gaps. Additionally, another key challenge lies in the reliance on automated evaluation metrics such as ROUGE. While useful for measuring surface-level similarities between generated and reference texts, these metrics may fail to account for deeper aspects of text quality, such as coherence, logical flow, and factual correctness.

Future work should focus on balancing the language data, expanding coverage of specialized topics, and developing evaluation frameworks that better align with human judgment while improving model robustness to reduce factual inconsistencies.

\bibliography{acl_latex}

\appendix
\clearpage
\section{Prompt Examples}
\label{sec:appendixA}
\subsection{Detailed Description of Prompt Examples}
The table \ref{tab:prompt_example} presents three examples of \textbf{prompt templates} designed for long-form question-answer generation based on provided abstracts. These templates help structure the interaction between the question and abstract, ensuring diverse and high-quality responses.

\textbf{Structure and Functionality:} Each prompt is structured to include one or more abstract references (denoted as \textit{\{Abstract[1]...Abstract[2]\}}) and a dynamically generated question (denoted as \textit{\{Question\}}). These prompts serve to guide the model in providing detailed, relevant answers while ensuring that the content aligns with the source material.

\paragraph{Prompt Diversity:} To ensure that the generated content covers a wide range of information points and avoids repetition, multiple prompt templates are provided. For instance:
\begin{itemize}
    \item \textbf{Prompt 1} requests an answer based on the references, emphasizing completeness.
    \item \textbf{Prompt 2} adds a directive that the model must answer, reinforcing the requirement for response.
    \item \textbf{Prompt 3} simplifies the directive but maintains the expectation of a detailed answer.
\end{itemize}

\paragraph{Customization for Content Length:}
The prompts are adaptable for abstracts of different lengths. By adjusting the amount of information in \texttt{\{Abstract[1]...Abstract[2]\}}, the system ensures that both short and long references are effectively processed.

\paragraph{Advantages of Diverse Prompting:}
\begin{itemize}
    \item \textbf{Increased Diversity:} Multiple prompts ensure a broader coverage of information, reducing redundancy and improving the model's capability to generate varied responses.
    \item \textbf{Improved Robustness:} These prompts expose the model to different question structures and contexts, improving its ability to respond appropriately across diverse scenarios.
    \item \textbf{Enhanced Alignment:} By consistently referring to abstracts and generating questions dynamically, the prompts ensure that the answers remain aligned with the source information, enhancing both factual accuracy and relevance.
\end{itemize}

\subsection{Example of Generated Prompts}

\begin{itemize}
    \item \textbf{Example 1:}
    \begin{itemize}
        \item Abstract(s): A collection of summaries from referenced papers.
        \item Question: \textit{"How did AlphaGo's match against Fan Hui mark a turning point in its history?"}
    \end{itemize}
    \textbf{Prompt:}
    Based on the provided references, answer the following questions. 
    Please provide detailed answers with a minimum of 300 words:
    Abstract[1]...Abstract[2]... 
    Question: How did AlphaGo's match against Fan Hui mark a turning point in its history?
    \item 
    \textbf{Example 2:}
    \begin{itemize}
        \item Abstract(s): Abstracts related to a medical breakthrough.
        \item Question: \textit{"What were the key factors that led to the success of the new treatment?"}
    \end{itemize}
    \textbf{Prompt:}
    You cannot refuse to answer the question. Please refer to the following information:
    Abstract[1]...Abstract[2]... 
    Answer the following questions. Please provide detailed answers with a minimum of 300 words: 
    What were the key factors that led to the success of the new treatment?
\end{itemize}
\begin{table*}[]
\centering
\footnotesize
\renewcommand{\arraystretch}{1.5} 
\setlength{\tabcolsep}{12pt} 
\begin{tabular}{m{2.5cm}<{\centering}||m{10cm}<{\centering}} 
\hline\hline 
\textbf{Prompt} & \textbf{Prompt Text} \\ 
\hline\hline 
\textit{Prompt 1} & \texttt{{Abstract[1]...Abstract[2]......}}\textbackslash n Answer the following questions based on the provided references. Please provide detailed answers with a minimum of 300 words: \textbf{\{Question\}} \\
\hline
\textit{Prompt 2} & You cannot refuse to answer the question. Please refer to the following information:\textbackslash n\texttt{{Abstract[1]...Abstract[2]......}}\textbackslash n Answer the following questions. Please provide detailed answers with a minimum of 300 words: \textbf{\{Question\}} \\
\hline
\textit{Prompt 3} & Based on the provided references, answer the following questions. Please provide detailed answers with a minimum of 300 words:\textbackslash n\texttt{{Abstract[1]...Abstract[2]......}}\textbackslash n Question: \textbf{\{Question\}} \\
\hline\hline 
\end{tabular}
\caption{Prompt example (Using a paragraph length of 200-400 characters as an example.)}
\label{tab:prompt_example}
\end{table*}
These examples demonstrate how diverse prompting enhances content generation, ensuring a balance between coverage and specificity in the answers.


\section{Hallucination Detection Algorithm for Citation Reliability}
\label{sec:appendix B}

The \textbf{Hallucination Detection Algorithm for Citation Reliability (HDACR)} is a crucial component designed to ensure the factual consistency between generated content and reference material in long-form article generation tasks. This algorithm specifically targets the identification of hallucinations, which are instances where the generated content introduces false or unverifiable information that is not present in the reference material. Below is a detailed explanation of the steps involved in the HDACR algorithm, as presented in Algorithm~\ref{alg:HDACR}.
\subsection{Step-by-Step Breakdown of HDACR}
\textbf{Initialization:}
The algorithm starts by initializing two sets: the \textit{Reference Content Entity Set} ($E_r$) and the \textit{Generated Content Entity Set} ($E_g$), which will store the extracted entities from the reference content ($R$) and the generated content ($G$), respectively. An empty \textit{Hallucination Detection Result} ($H$) is also initialized to store the final outcome of the detection process.

\textbf{Entity Extraction:}
Next, the algorithm extracts entities from both the reference content ($R$) and the generated content ($G$). For each model $M$ in the preDeFined model set, the entities from the reference content are extracted and added to $E_r$. Similarly, entities from the generated content are extracted and added to $E_g$. This entity extraction process ensures that both texts (reference and generated) are represented by the key entities they contain, facilitating accurate comparison.

\textbf{Matching Score Calculation:}
Once the entities are extracted, the algorithm proceeds to calculate a \textit{matching score} for each entity in the generated content ($E_g$). If an entity from the generated content has a direct match (hard match) in the reference content ($E_r$), a perfect score of 1.0 is assigned. If no hard match is found, the algorithm performs a \textit{soft matching} using a combination of Sentence-BERT (Semantic-BERT) and BM25 scoring methods. The weighted score is computed by averaging the Sentence-BERT and BM25 scores for each entity. The final score is stored in the matching score list $\gamma$.

\textbf{Hallucination Detection:}
The hallucination detection decision is based on the matching scores. If any entity in the generated content has a matching score $\gamma_i$ lower than a preDeFined threshold (0.6 in this case), the algorithm concludes that the generated content contains a hallucination. Otherwise, the content is deemed free of hallucinations.

\textbf{Output:}
Finally, the algorithm outputs the hallucination detection result ($H$), along with the generated content ($G$), the reference content ($R$), and a detailed list of unverifiable entities and their positions within the text. This output provides a comprehensive report on the factual consistency of the generated content, helping to ensure its reliability.

\begin{algorithm*}[] 
\SetAlgoLined
\caption{Hallucination Detection Algorithm for Citation Reliability (HDACR)}
\KwIn{Generated content $G$ and Reference content $R$}
\KwOut{Hallucination detection result $H$}
\BlankLine
\textbf{Initialization}: Reference content entity set $E_r \gets \emptyset$, Generated content entity set $E_g \gets \emptyset$, \hspace{2em} Hallucination detection result $H \gets$ empty\;
\textbf{Extract Reference Content Entities}: For each model $M$ in the model set, extract entities  $E_{rM}$ from the reference content $R_r$, add $E_{rM}$ to $E_r$\;
\textbf{Extract Generated Content Entities}: For each model $M$ in the model set, extract entitie$E_{gM}$ from the generated content $G$, add $E_{gM}$ to $E_g$\;
\textbf{Calculate Matching Scores}: Initialize the matching score list $\gamma \gets []$. For each entity $e$ in $E_g$:
   \begin{itemize}
     \item If $e$ matches hard ($e \in E_r$), then $\gamma \gets 1.0$\;
     \item Otherwise, perform soft matching, calculate Sentence-BERT score $\gamma_{SBER T}$ and BM25 score $\gamma_{BM25}$, calculate weighted score $\gamma \gets \frac{\gamma_{SBER T} + \gamma_{BM25}}{2}$, add $\gamma$ to the matching score list.
   \end{itemize}
\textbf{Determine the Presence of Hallucinations}: If $\exists \gamma_i < 0.6$, mark hallucination detection result: $H \gets$ Hallucination present; otherwise, mark hallucination detection result $H \gets$ No hallucination
\textbf{Output Results}: Output generated content $G$, hallucination detection result $H$, reference content $R$, and unverifiable entities and their positions in the text\;
\label{alg:HDACR}
\end{algorithm*}

The following is the pseudocode for the HDACR algorithm. The HDACR algorithm plays a critical role in improving the accuracy and reliability of long-form generated content. By carefully comparing entities between the reference and generated content, it ensures that no extraneous or fabricated information is introduced. This makes it particularly valuable in domains where factual consistency is crucial, such as scientific or historical document generation. The combination of both hard and soft matching provides flexibility in detecting minor semantic variations while maintaining rigorous detection of factual hallucinations. The output report generated by HDACR offers transparency into the hallucination detection process, making it an essential tool for validating the quality of generated content.

\section{Dataset Statistics and Distribution}
\label{sec:appendixDatasetStats}
We randomly selected 2,286 outline data samples and 3,625 question-answer pairs, with an equal split between Chinese and English, from DeFine dataset to fine-tune the Qwen2-7b-Instruct model~\citep{yang2024qwen2technicalreport} for long-form article generation tasks in order to validate the effectiveness of our dataset.
To further enhance the model's capabilities in information extraction and compression tasks, we trained the BGE-m3 relation extraction model using abstract set data. This model is designed to extract key information from references and condense full-text content into summaries through precise relation extraction and compression.

From the remaining data, 270 outline samples and 270 QA pairs were reserved to evaluate the outline generation and long-article generation performance of the fine-tuned model. The fine-tuned model is referred to as \textbf{\textit{Qwen2-7b-Scribe}}. To validate the effectiveness of our methods, we selected 230 distinct topics from various domains, including 130 topics in Chinese and 100 in English, to generate full-length articles.

Table~\ref{tab:dataset_stats} provides comprehensive statistics on the various components of the dataset used in this study. Each category is further divided into English (eng.) and Chinese (ch.) data, with columns detailing the number of examples in the test set, training set, and the total number of examples available.

Notably, the English data contains a significantly larger number of samples compared to the Chinese data. This discrepancy arises from the larger volume of English-language Wikipedia entries; it has a far richer and more extensive collection of articles than its Chinese counterpart, with a greater number of contributors and editors actively engaged in expanding and updating content. According to recent studies, English Wikipedia hosts over 6 million articles, while Chinese Wikipedia has just over 1 million articles ~\citep{wiki2024}. This has resulted in a more substantial dataset for English, leading to the selection of a representative subset to avoid redundancy and overfitting in training. By doing so, we aimed to ensure that the training process remains efficient while still providing the model with sufficient variability and representation from the dataset.

Furthermore, the varying depth and detail within entries also contribute to the richness of the English dataset. Articles on complex topics often include extensive references and citations, which enhance the dataset's quality and comprehensiveness. In contrast, the Chinese entries, while valuable, may not always match this level of detail, particularly in less commonly covered subjects. This situation necessitates a careful curation of data to maintain a balanced representation in the model's training set, ensuring that both languages are adequately represented in the resulting outputs.

\begin{figure*}[!htbp]
  \centering
  \includegraphics[width=\textwidth]{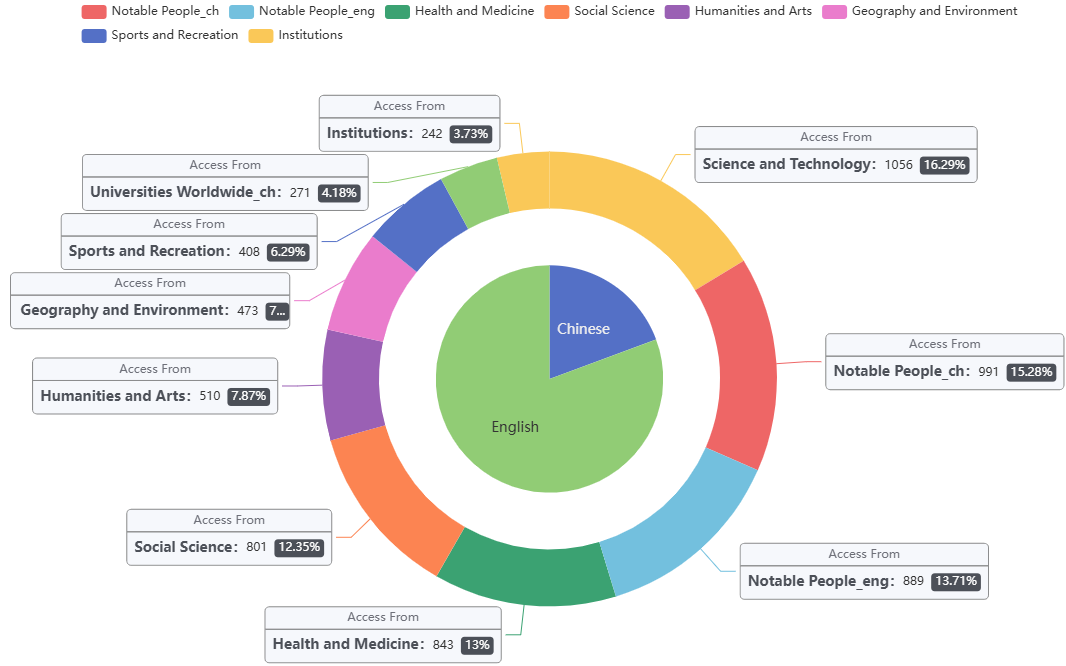}  
  \caption{This figure illustrates the access distribution of QA data topics in the dataset. The categories ending with \textbf{ch} represent Chinese data, while all other categories are in English. The percentages indicate the proportion of accesses from each topic, highlighting the diversity in the dataset.}
  \label{fig:QA data distribution}
\end{figure*}

\begin{figure*}[!htbp]
  \centering
  \includegraphics[width=\textwidth]{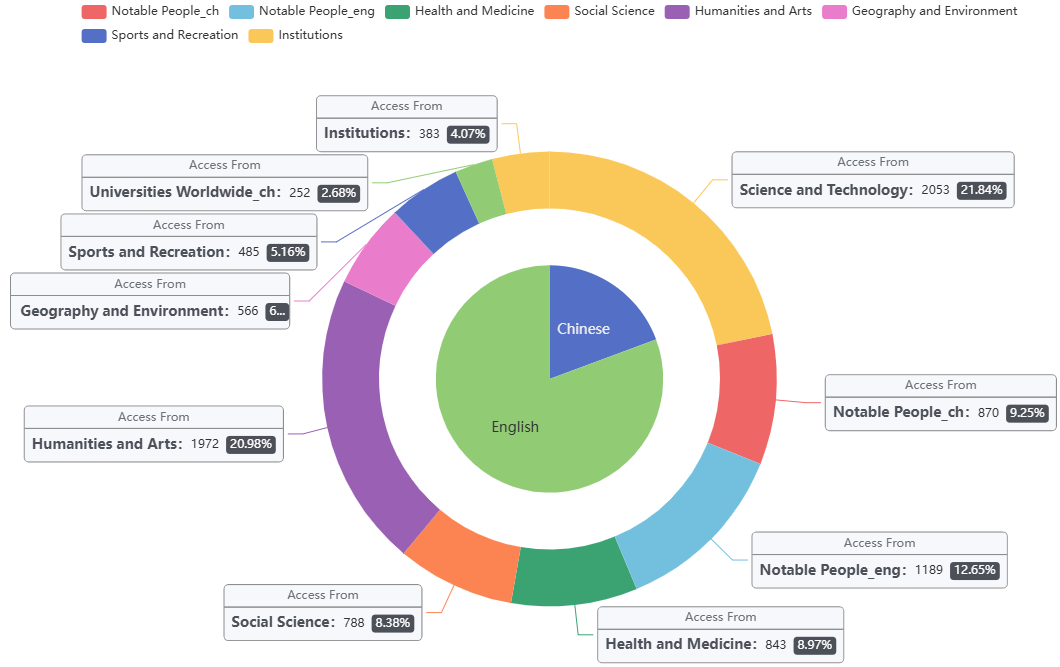}  
  \caption{This figure illustrates the access distribution of Abstract set data topics in the dataset. The categories ending with \textbf{ch} represent Chinese data, while all other categories are in English. The percentages indicate the proportion of accesses from each topic, highlighting the diversity in the dataset.}
  \label{fig:Abstract data distribution}
\end{figure*}

The distribution of various data types is shown in Figure~\ref{fig:QA data distribution} and Figure~\ref{fig:Abstract data distribution}. Figure~\ref{fig:QA data distribution} illustrates the access distribution of QA data topics in the dataset, while Figure~\ref{fig:Abstract data distribution} provides similar insights for Abstract data topics. The categories ending with \textbf{ch} represent Chinese data, while all other categories correspond to English data. The percentages displayed in the figures indicate the proportion of accesses from each topic, highlighting the dataset's diversity. This rich distribution is essential for training models capable of understanding and generating content across a variety of topics and styles, further enhancing the robustness and applicability of our findings.

\begin{table}[]
\centering
\small
\renewcommand{\arraystretch}{1.2} 
\setlength{\tabcolsep}{8pt} 
\begin{tabular}{|c|c|c|c|}
\hline
\textbf{Data Type} & \textbf{Test.} & \textbf{Train.} & \textbf{All} \\ 
\hline
Outline data eng. & 120 & 1026 & 1226 \\ 
\hline
Outline data ch. & 150 & 941 & 50787 \\ 
\hline
Abstract Set Data eng. & \textbackslash & 200 & 8525 \\ 
\hline
Abstract Set Data ch. & \textbackslash & 200 & 1122 \\ 
\hline
QA data eng. & 133 & 1026 & 5244 \\ 
\hline
QA data ch. & 120 & 1138 & 1258 \\ 
\hline
\end{tabular}
\caption{This table provides detailed information about our dataset. The dataset is categorized into Outline data, Abstract Set data, and QA data, with separate columns for English (eng.) and Chinese (ch.). The backslash (`\textbackslash`) denotes that no test data is available for the Abstract Set in either English or Chinese.}
\label{tab:dataset_stats}
\end{table}

\section{Case Study}

In this appendix, we present examples of the results generated by different models used in our experiments. We provide outlines, QA data, and full generated articles for several topics.

\subsection{Outline Generation Examples}
Here is the section where we present the outlines generated by different models for the same task and compare them with the correct answers. The task is to generate a directory for the article topic "\begin{CJK}{UTF8}{gbsn}首尔大学\end{CJK}" and "John Gribbel." Below, we list the generated results from each model alongside the correct answers.

\subsection{Chinese Outline Generation}

\textbf{Instruction}: \begin{CJK}{UTF8}{gbsn}请根据‘首尔大学’这一主题，生成一篇长篇文章的目录结构。目录应包括多个主要部分和其子部分，以确保涵盖主题的各个方面。\end{CJK}

\subsubsection{Original Outline}
\begin{CJK}{UTF8}{gbsn}
\begin{verbatim}
1. 摘要
2. 正文
    2.1 历史
        2.1.1 早期历史
        2.1.2 国立首尔大学时期
        2.1.3 冠岳校区
        2.1.4 莲建校区
        2.1.5 平昌校区
    2.2 学生与校友
    2.3 大学排名
        2.3.1 QS世界大学排名
        2.3.2 泰晤士高等教育世界大学排名
        2.3.3 软科世界大学学术排名
    2.4 交通
        2.4.1 校本部
        2.4.2 莲建校区
        2.4.3 平昌校区
    2.5 参看
    2.6 外部连结

\end{verbatim}
\end{CJK}
\subsubsection{Llama3 Output}
\begin{CJK}{UTF8}{gbsn}
\begin{verbatim}
    - 基本信息
    - 历史沿革
    - 学术研究
    - 院系设置
    - 校园设施
    - 国际交流
    - 名人校友
    - 社会影响
\end{verbatim}
\end{CJK}

\subsubsection{Qwen1.5-7b-chat Output}
\begin{CJK}{UTF8}{gbsn}
\begin{verbatim}
1. 基本信息
   - 全称
   - 成立时间
   - 校址位置
2. 概述
   - 历史背景
   - 地位与排名
3. 组织结构
   - 校级管理层
   - 学院/学部
4. 教育体系
   - 本科教育
   - 研究生课程
5. 校园文化
   - 校园生活
   - 学生社团
6. 未来发展
    - 发展规划
    - 教育改革
7. 链接
    - 官方网站
    - 联系方式
\end{verbatim}
\end{CJK}

\subsubsection{Qwen2-7b-Instruct Output}
\begin{CJK}{UTF8}{gbsn}
\begin{verbatim}
一、简介
   - 学校概况
   - 校训与理念
二、学术与研究
   - 学科设置
   - 研究机构与中心
   - 学术成就与奖项
三、校园与设施
   - 校园布局与环境
   - 学生生活设施
四、国际交流
   - 国际合作与交流项目
五、社会与社区
   - 社区服务与社会责任
   - 校友捐赠与支持
六、未来展望
   - 学校发展规划
   - 面向未来的教育理念与目标
\end{verbatim}
\end{CJK}

\subsubsection{GPT-3.5}
\begin{CJK}{UTF8}{gbsn}
\begin{verbatim}
I. 首尔大学的历史
II. 首尔大学的地理位置
III. 首尔大学的校园建筑
IV. 首尔大学的学术声誉
V. 首尔大学的院系设置
VI. 首尔大学的师资队伍
VII. 首尔大学的学生组成
VIII. 首尔大学的校园文化
IX. 首尔大学的研究成果
X. 首尔大学的国际交流
XI. 首尔大学的校友影响
XII. 首尔大学的知名校友
XIII. 首尔大学的主要校园活动
XIV. 首尔大学的校园风景
XV. 首尔大学的校园发展规划
XVI. 首尔大学的学术亮点
XVII. 首尔大学的社会服务
XVIII. 首尔大学的校园科技
XIX. 首尔大学的未来展望
XX. 首尔大学的特色建筑
\end{verbatim}
\end{CJK}

\subsubsection{Qwen2-7b-Scribe Output}
\begin{CJK}{UTF8}{gbsn}
\begin{verbatim}
1. 摘要
2. 正文
    2.1 学校概况
        2.1.1 校史与创办理念
        2.1.2 校区分布
        2.1.3 校训与文化特色
    2.2 学术与研究
        2.2.1 主要学科及研究方向
        2.2.2 重点研究机构与实验室
        2.2.3 学术影响力与排名
        2.2.4 知名学术奖项与科研成果
    2.3 校园设施与环境
        2.3.1 校园建筑与文化空间
        2.3.2 学术资源：图书馆与在线资源
        2.3.3 校园交通
    2.4 教育体系
        2.4.1 本科教育与课程设置
        2.4.2 研究生及博士后项目
        2.4.3 国际课程与交换生项目
    2.5 国际化与全球影响
        2.5.1 国际合作与学术交流
        2.5.2 校外国际学术活动
        2.5.3 国际学生及教师组成
    2.6 校友与社会贡献
        2.6.1 校友在国内外的影响力
        2.6.2 校友捐赠与社会责任
    2.7 校园文化与学生生活
        2.7.1 学生组织与文化活动
        2.7.2 校园多样性与包容性
    2.8 挑战与机遇
    2.9 未来展望与发展
\end{verbatim}
\end{CJK}

\subsection{John Gribbel}

\textbf{Instruction}: Please generate a directory structure for a long-form article based on the theme of ‘John Gribbel.’ The directory should include multiple main sections and subsections to ensure all aspects of the topic are covered.

Below we present the outlines generated by different models compared to the correct answer.

\subsubsection{Original Outline}
\begin{verbatim}
1. Abstract
2. Main text
   2.1 Early life and education
   2.2 Gas, electricity and 
          other business interests
   2.3 Residences
   2.4 Litigation
   2.5 Publishing
   2.6 Politics
   2.7 History and the 
          Glenriddell Manuscripts
   2.8 Marriage
\end{verbatim}

\subsubsection{Llama3 Output}
\begin{verbatim}
- Biography
- Early Life
- Career Highlights
- Notable Works
- Contributions to the Field
- Awards and Recognition
- Personal Life
- Legacy
\end{verbatim}

\subsubsection{Qwen1.5-7b Output}
\begin{verbatim}
1. Biographical Overview
2. Full Name: John Gribbel
3. Birth and Early Life
4. Birthplace
5. Date of Birth
6. Education and Background
7. Academic Qualifications
8. Professional Career
9. Notable Works
10. Influence on Others
\end{verbatim}

\subsubsection{Qwen2-7b-Instruct Output}
\begin{verbatim}
I. Introduction
   A. Brief overview of John Gribbel
II. Early Life and Education
   A. Birth and family background
III. Career and Professional Contributions
IV. Research and Publications
   A. Collaborations with other experts
V. Personal Life and Interests
   A. Family, hobbies, and personal passions
VI. Conclusion
\end{verbatim}

\subsubsection{GPT-3.5}
\begin{verbatim}
Early Life and Background
Education and Academic Achievements
Research and Discoveries
Notable Publications
Awards and Recognition
Influence on Future Scientists
Personal Life and Family
Collaborations and Partnerships
Controversies and Criticisms
Public Perception and Media Coverage
Enduring Contributions and Significance
\end{verbatim}

\subsubsection{Qwen2-7b-Scribe Output}
\begin{verbatim}
1. Abstract
2. Main text
    2.1 Early Life and Education
        2.1.1 Birth and Family Background
        2.1.2 Academic Achievements
    2.2 Career
        2.2.1 Early Career
        2.2.2 Major Contributions
        2.2.3 Later Career
    2.3 Personal Life
        2.3.1 Family and Relationships
        2.3.2 Personal Interests and Hobbies
    2.4 Awards and Honors
    2.5 Selected Publications
        2.5.1 Books and Major Works
        2.5.2 Articles and Research Papers
    2.6 External Links
    2.6.1 Relevant Biographical Websites
    2.6.2 Collaborative Projects and Initiatives
\end{verbatim}

\subsection{QA Data Generation Examples}
Here is the section where we present the paragraphs generated by Qwen2-7b-Scribe in three different pipelines for the same task and compare them with the correct answers. The task is to generate an article paragraph for the topic "\begin{CJK}{UTF8}{gbsn}《中国季刊》\end{CJK}" and "John Gribbel." 

\noindent\textbf{Topic}:  \begin{CJK}{UTF8}{gbsn}《中国季刊》\end{CJK}\\
\textbf{Question}:  \begin{CJK}{UTF8}{gbsn}《中国季刊》的历史和发展的主题是什么？\end{CJK}\\
\textbf{Original Answer}: \begin{CJK}{UTF8}{gbsn}曾任哈佛大学政治系教授的马若德以研究中国文化大革命史而闻名，他于1960年在伦敦英国国际关系研究所创办了《中国季刊》，它是国外第一份专门研究当代中国的学术刊物，起初由冷战时期美国中央情报局的“自由文化基金会”（Congress for Cultural Freedom,CCF）出资运营。1967年自由文化基金会解散，《中国季刊》移交给伦敦大学亚非学院，这本杂志在客观上推动了西方中国研究从政策研究转变向学术研究。自此，一些与传统的对中国问题的研究方法和方式不同的诸如比较研究、跨国合作研究、多学科联合研究等，也由这本杂志最早出现并陆续在其它刊物上流行。目前，《中国季刊》是西方学者发表与中国近现代历史和中华人民共和国史相关的学术研究的重要刊物，也是国际上研究中国相关问题的杂志中影响因子最高的杂志（1.0左右，2006年1.216）。\end{CJK}\\

\subsubsection{Web Retrieval Baseline}
\textbf{Generated Answer By Qwen2-7b-Scribe}:  \begin{CJK}{UTF8}{gbsn}《中国季刊》（The China Quarterly）作为一本国际著名的学术期刊，其历史和发展的主题经历了显著的变化。自1960年创刊以来，该刊最初由国际著名的中国问题专家罗德里克·麦克法夸尔（Roderick MacFarquhar）创立，旨在反映铁幕背后中国各领域的状况，深入理解中共历史和管治逻辑。初期，刊物主要聚焦于政治学，研究与美国外交政策相关的课题，不可避免地带有较强烈的意识形态倾向。

然而，随着时间的推移，尤其是进入21世纪后，《中国季刊》的编辑理念和内容范围发生了显著变化。普林格尔指出，中国学者越来越多地参与到有关中国的研究中，并且用英文发表研究成果，这标志着中国研究领域的一个重大转变。这不仅反映了中国学者在国际学术舞台上的日益重要地位，也体现了中国研究的多元化和深入化。

此外，随着全球学术交流的加强，越来越多的学者有机会接触中国，参与中国的发展。《中国季刊》在这一过程中发挥了主导作用，推动了学术界对中国的全面认识。例如，希娜·切斯特纳特·格雷腾斯（Sheena Chestnut Greitens）和罗里·特鲁克斯（Rory Truex）的文章揭示了非中国学者与中国“国保”（即国内安全保卫）人员之间的交流，以及被请去“喝茶”（指被约谈）的情况，这表明“喝茶”并不像人们想象地那么普遍，但依旧是个问题。

值得注意的是，《中国季刊》在2009年从一年四期扩展到一年六期，并在2021年和2022年连续八期扩大版面，消化了当时积压的一到两年的待刊文稿。这不仅体现了《中国季刊》在学术资源利用上的高效，也反映了其在学术界的地位和影响力。

综上所述，《中国季刊》的历史和发展的主题经历了从政治学向历史学、社会学、人类学和文学等多学科的转变，反映了中国研究领域的多元化和深入化。这一转变不仅得益于中国学者的积极参与，也得益于全球学术交流的加强，体现了《中国季刊》在推动中国研究方面的重要作用。"\end{CJK}\\

\subsubsection{Local Retrieval Baseline}
\textbf{Generated Answer By Qwen2-7b-Scribe}:  \begin{CJK}{UTF8}{gbsn}“中国季刊”（The China Quarterly）自1960年创刊以来，始终致力于探讨中国历史、政治、社会、文化等多方面的议题，尤其聚焦于中华人民共和国的各个方面和中国共产党的历史。麦克法夸尔，作为该刊的第一任主编（1960—1968），在创刊之初就明确了刊物的宗旨，即不仅要反映中国国内的政治、经济、社会等动态，还要关注当代台湾问题和海外华人的文章，以全面展现中国的发展面貌。在麦克法夸尔的领导下，该刊逐渐成为海外中国研究领域影响力最大的刊物之一，不仅能够收到投稿，还吸引了众多学者和研究人员的关注。

随着时代的发展，中国研究领域发生了显著变化。中国学者越来越多地参与到有关中国的研究中，并且用英文发表研究成果，这标志着中国研究的国际化进程加速。此外，中国季刊也面临着新的挑战，尤其是在安全和保障问题上，需要找到创新的研究方法，并为匿名研究提供安全的环境。这不仅要求研究者在道德上和方法论上给予更多关怀，还意味着需要在学术自由与国家安全之间找到平衡点。

在历史人物的塑造方面，无论是影视作品还是学术研究，都需要深入了解历史大背景、人物性格以及当时环境下对人物性格的影响。历史人物的性格是塑造其形象的关键，而了解这些因素有助于更准确地还原历史人物的真实面貌，避免过度美化或丑化。在学术研究中，深入探讨历史人物的性格及其在特定历史时期的行为逻辑，对于揭示历史事件的深层原因和人物之间的复杂关系具有重要意义。

综上所述，“中国季刊”作为一本专注于中国研究的学术期刊，其历史和发展的主题涵盖了中国各领域的最新研究成果，反映了中国学者在国际学术舞台上的日益活跃，同时也面临着如何在学术自由与国家安全之间寻求平衡的挑战。在历史人物的塑造上，深入研究其性格和行为逻辑，对于理解历史事件和人物形象具有不可或缺的作用。\end{CJK}\\

\subsubsection{Grounded Reference Baseline}
\textbf{Generated Answer By Qwen2-7b-Scribe}:  \begin{CJK}{UTF8}{gbsn}“中国季刊”（The China Quarterly）是1960年在英国伦敦创立的学术期刊，由美国学者麦克法夸尔（John King Fairbank）担任首任主编。该期刊的创立标志着西方中国学从政策研究向学术研究的转变，麦克法夸尔在任期间积极组织学术活动，推动了西方当代中国研究的发展。然而，由于麦克法夸尔的逝世，该期刊在1968年停刊。1973年，美国学者费正清（John King Fairbank）接任主编，重新启动了该期刊，并将其定位为“中国研究的国际性论坛”。费正清在任期间，邀请了众多中国问题专家投稿，包括中国学者，如李慎之、黄仁宇等。该期刊的复刊，标志着西方中国学研究的复兴，成为全球中国研究的重要平台。然而，由于费正清的逝世，该期刊在1988年再次停刊。此后，该期刊由美国学者罗伯特·库恩（Robert E. Carter）担任主编，继续出版。该期刊的历史和发展的主题，是西方中国学从政策研究向学术研究的转变，以及全球中国研究的发展和方向。\end{CJK}\\

\noindent\textbf{Topic}: Medical Prescription \\
\textbf{Question}:  What is the importance of legibility in the writing of medical prescriptions, and how can healthcare professionals ensure that their prescriptions are easily readable and understandable to patients and pharmacists?\\
\textbf{Original Answer}: Doctors' handwriting is a reference to the stereotypically illegible handwriting of some medical practitioners, which sometimes causes errors in dispensing. In the US, illegible handwriting has been indirectly responsible for at least 7,000 deaths annually.There are several theories about the causes of this phenomenon. Some sources say the extreme amount of writing doctors employ during training and at work leads to bad handwriting,[better source needed]whereas others claim that doctors neglect proper handwriting due to medical documents being intended to be read solely by medical professionals, not patients.[better source needed]Others simply classify the handwriting of doctors as a handwriting style.The issue may also have a historical origin, as physicians from Europe-influenced schools have historically used Latin words and abbreviations to convey prescriptions;many of the abbreviations are still widely used in the modern day and could be a source of confusion.Some jurisdictions have legislatively required prescriptions to be legible—Florida specifies \textit{"legibly printed or typed"}—and the Institute for Safe Medication Practices advocated the elimination of handwritten prescriptions altogether.There have been numerous devices designed to electronically read the handwriting of doctors, including electronic character recognition,keyword spotters,and \textit{"postprocessing approaches"},though the gradual shift to electronic health records and electronic prescriptions may alleviate the need for handwritten prescriptions altogether.In Britain's NHS, remaining paper prescriptions are almost invariably computer printed, and electronic (rather than paper) communication between surgery and pharmacy is increasingly the norm.

\subsubsection{Web Retrieval Baseline}
\textbf{Generated Answer By Qwen2-7b-Scribe}: Legibility in medical prescriptions is of paramount importance for several reasons. Firstly, it ensures patient safety by preventing medication errors, which can occur when pharmacists misinterpret illegible or ambiguous instructions. Secondly, it facilitates efficient communication between healthcare providers, as clear prescriptions allow for quicker and more accurate dispensing of medications. Lastly, it respects the professional duties of prescribers to provide legible and complete orders, which is a legal requirement in many jurisdictions.

To ensure that their prescriptions are easily readable and understandable, healthcare professionals can take several steps:

n1.\textbf{Use of Standardized Formats}: Adhering to established formats for prescriptions, such as those recommended by the World Health Organization (WHO), can help standardize the layout and structure of prescriptions, making them more uniform and easier to read.

n2.\textbf{Clear and Concise Language}: Avoiding medical jargon and using clear, concise language can prevent misunderstandings. It is also helpful to include the patient's name, date, and the prescriber's signature to provide context and authenticity.

n3.\textbf{Legibility of Handwriting}: Practicing good handwriting skills is crucial, especially for those who write prescriptions by hand. Using a consistent and clear handwriting style can significantly improve readability.

n4.\textbf{Use of Authorized Abbreviations:} While abbreviations can save time, they must be used correctly and understood by all involved parties. Misinterpretation of abbreviations can lead to errors, so it's important to use them judiciously and ensure that all parties are familiar with the abbreviations being used.

n5.\textbf{Complete Information:} Including all necessary information such as the patient's name, date, time, medication name, strength, quantity, and route of administration ensures that the prescription is comprehensive and reduces the risk of errors.

n6.\textbf{Review and Feedback:} Regularly reviewing prescriptions for legibility and completeness, and seeking feedback from peers or pharmacists, can help identify areas for improvement and ensure that prescriptions meet the required standards.

n7.\textbf{Education and Training:} Healthcare professionals should receive ongoing education and training on the importance of legibility and completeness in prescriptions. This can include workshops, seminars, and online resources that focus on best practices in prescription writing.

n8.\textbf{Use of Technology:} Employing electronic health records (EHRs) and digital prescription systems can enhance legibility and reduce errors. These systems often have built-in checks for completeness and legibility, and they can also help in tracking and managing medication orders more efficiently.

By implementing these strategies, healthcare professionals can significantly improve the legibility of their prescriptions, thereby contributing to safer and more effective patient care.

\subsubsection{Local Retrieval Baseline}
\textbf{Generated Answer By Qwen2-7b-Scribe}: Legibility in the writing of medical prescriptions is of paramount importance for ensuring patient safety and effective medication management. A clear and legible prescription minimizes the risk of errors, such as misinterpretation of dosages or incorrect medication administration, which can lead to adverse drug events. Healthcare professionals can ensure that their prescriptions are easily readable and understandable by following several best practices:

n1.\textbf{Use of Standardized Abbreviations:} Adhere to universally accepted medical abbreviations and symbols to avoid confusion. For instance, \"q.d.\" for \"every day\" or \"p.o.\" for \"per os\" (by mouth) should be used consistently.

n2.\textbf{Avoidance of Acronyms:} While some acronyms are standard in medical practice, using them without explanation can lead to misunderstandings, especially for patients who are not familiar with medical terminology. Always spell out acronyms the first time they are used in a prescription.

n3.\textbf{Clear and Concise Instructions:} Write instructions in a straightforward manner, avoiding medical jargon that might not be understood by patients or pharmacists. For example, instead of writing \"take with food,\" specify \"with meals\" or \"before meals.

n4.\textbf{Use of Proper Units of Measurement:} Always include the unit of measurement (e.g., mg, mL) to avoid confusion. Ensure that the unit is consistent with the medication’s standard packaging.

n5.\textbf{Dosage and Frequency:} Clearly state the dosage and how often it should be administered. For instance, ``take 1 tablet twice daily'' is more understandable than ``take 1 tablet q.d.''

n6.\textbf{Prescriber’s Signature:} Include a clear signature or initials to authenticate the prescription. This is crucial for legal and accountability purposes.

n7.\textbf{Prescription Number and Date:} Include a unique prescription number and the date of issuance. This helps in tracking the prescription and managing inventory.

n8.\textbf{Dosage Forms and Route of Administration:} Specify the dosage form (e.g., tablet, capsule) and the route of administration (e.g., orally, intravenously) to ensure the correct medication is administered in the right way.

n9.\textbf{Dosage Instructions for Children:} For pediatric patients, provide clear instructions on how to measure liquid medications or how to adjust dosages based on weight or age.

n10.\textbf{Prescription Review:} Before finalizing a prescription, review it for clarity and completeness. It can be helpful to have a colleague or pharmacist review the prescription to catch any potential errors.

By following these guidelines, healthcare professionals can significantly reduce the likelihood of prescription errors, thereby enhancing patient safety and the overall effectiveness of medication management.

\subsubsection{Grounded Reference Baseline}
\textbf{Generated Answer By Qwen2-7b-Scribe}: The importance of legibility in the writing of medical prescriptions is paramount to ensure that patients receive the correct medication and dosage. Inaccurate or illegible prescriptions can lead to medication errors, which can have serious consequences for patient health. To ensure that their prescriptions are easily readable and understandable, healthcare professionals should write in a clear and consistent manner, using legible handwriting or electronic medical records. Additionally, healthcare professionals should double-check their prescriptions for accuracy and completeness before submitting them to a pharmacist. Pharmacists also have a responsibility to ensure that prescriptions are legible and to clarify any ambiguities with the prescriber. In some jurisdictions, such as the United States, there are laws and regulations that require prescriptions to be legibly printed or typed and to contain specific information, such as the name of the prescribing practitioner, the name and strength of the drug prescribed, the quantity of the drug prescribed, and the directions for use of the drug.

\subsection{Full Article Generation Examples}
In the spirit of rigorous academic comparison, we select an entry \textit{Taylor Hawkins} from FreshWiki dataset\citep{shao2024assisting}, same as STORM\citep{} presented in their paper's appendix. We use our best performed pipeline: Web Retrieval pipeline, and set the generated model as Qwen2-7b-Scribe. we provide a comprehensive article generated by our model, showcasing its ability to produce detailed, long-form text on complex subjects. In addition, we include the long-form article on the same topic generated in the STORM paper\citep{shao2024assisting} in the appendix of our document for a direct comparison with our results.

\onecolumn
\renewcommand{\arraystretch}{1.7} %

\begin{longtable}{|p{1.0\textwidth}|}
\hline
\texttt{{\textbf{Taylor Hawkins}}} \footnotesize{\textit{(Article generated by our Web Retrieval Method)}} \\
\hline
\endfirsthead
\hline
\texttt{\small{Taylor Hawkins, the legendary drummer of the Foo Fighters, left an indelible mark on the world of rock music with his dynamic drumming and charismatic stage presence. Born in 1972 in Fort Worth, Texas, and raised in Laguna Beach, California, Hawkins' passion for music was ignited at a young age, culminating in a career that would influence generations of musicians. His journey began on the drums for Alanis Morissette before joining the Foo Fighters in 1997, where he contributed to the band's iconic sound through albums like "The Colour and the Shape" and "Sonic Highways."}} \\
\texttt{\small{Hawkins was not only known for his powerful drumming but also for his vocal talents and songwriting, which were showcased in tracks like "Cold Day in the Sun." Beyond the Foo Fighters, his musical explorations included his side project, Taylor Hawkins and the Coattail Riders, where he expressed his diverse musical tastes. Despite personal struggles, including a battle with substance abuse, Hawkins' resilience and dedication to his craft left a legacy of inspiration.}} \\
\texttt{\small{His unexpected death in 2022 shocked the music community, but his influence continues through the music he made and the lives he touched. Remembered for both his vibrant music and vibrant personality, Hawkins remains a symbol of perseverance and passion in the music industry.}} \\
\hline
\endhead
\hline
\endfoot
\hline
\endlastfoot

\texttt{\small{\textbf{\# Early Life and Career}}} \\
\texttt{\small{Taylor Hawkins, the late drummer of the Foo Fighters, had a life that was marked by both musical talent and personal struggles. Born Oliver Taylor Hawkins on February 17, 1972, in Fort Worth, Texas, Hawkins was raised in Laguna Beach, California, where he developed a passion for music. His early years were dedicated to honing his skills, and he immersed himself in the rich tapestry of musical styles that would later define his signature sound.}} \\
\texttt{\small{Hawkins' journey in music began with his first big break as the drummer for Grammy-winning artist Alanis Morissette. Joining her band in the mid-1990s, he toured extensively with her, contributing to the success of her iconic album Jagged Little Pill. This experience not only launched his career but also led him to meet his future bandmates in the Foo Fighters, as he replaced William Goldsmith in 1997.}} \\
\texttt{\small{His time with the Foo Fighters was marked by a whirlwind of activity, including the recording of their second album, The Colour and the Shape, which featured Hawkins' drumming and songwriting contributions. The album was a critical and commercial success, solidifying Hawkins' place in the music industry. He continued to work with the band, contributing to their subsequent albums and tours, which included the acclaimed Sonic Highways series and the 2022 zombie flick Studio 666.}} \\
\texttt{\small{Hawkins' musical skills extended beyond the drumset. He was also a talented singer and songwriter, as evidenced by his lead singing debut on an EP with a cover of Cream's "I Feel Free." He even played piano on the song "Summer's End" from Echoes, Silence, Patience \& Grace, showcasing his versatility as a musician.}} \\
\texttt{\small{Throughout his career, Hawkins recorded with various artists, including Coheed and Cambria, Eric Avery, Kerry Ellis, and Slash. His drumming and musical approach were influenced by Roger Taylor of Queen and Stewart Copeland of The Police, reflecting his diverse taste in music. He was also part of the SOS All-Stars with Roger Taylor of Queen and Chad Smith of Red Hot Chili Peppers, further highlighting his ability to collaborate with other musicians.}} \\
\texttt{\small{Hawkins' personal life was not without its challenges. He had a history of experimenting with drugs, which led to a near-fatal overdose in 2001. However, he was able to overcome this setback and live a clean lifestyle, enjoying a quieter life with his wife in California. Despite the distance between him and his bandmate Dave Grohl, their bond remained strong, reflecting the deep friendship that had developed over the years.}} \\
\texttt{\small{Hawkins' impact on the music world extended far beyond his drumming skills. He was a force of nature whose passion, talent, and spirit continued to inspire generations of musicians and fans alike. His legacy is not only in the music he created but also in the lives he touched and the memories he left behind.}} \\
\texttt{\small{In conclusion, Taylor Hawkins' early life and career were marked by dedication, determination, and a love for music. From his early days as a drummer for Alanis Morissette to his tenure with the Foo Fighters, he left an indelible mark on the music industry. His musical skills, versatility, and personal struggles made him a complex and fascinating figure, whose life and career continue to be celebrated and remembered.}} \\
\hline

\texttt{\small{\textbf{\# Role in the Foo Fighters}}} \\
\texttt{\small{Taylor Hawkins was the drummer for the American rock band Foo Fighters, a position he held from 1997 until his death in 2022. He joined the band in 1997, following the departure of original drummer William Goldsmith. Hawkins was known for his energetic and dynamic drumming style, which complemented the band's hard rock and alternative rock sound. He was also a talented singer and songwriter, contributing to the band's albums and even taking on lead vocals on occasion, such as on the song "Cold Day in the Sun" from the album In Your Honor. Hawkins' role in the band extended beyond the drums, as he was an integral part of the band's live performances, often engaging with the audience and adding a sense of humor and theatricality to the shows. He was also a respected musician in his own right, with side projects like Taylor Hawkins and the Coattail Riders, which allowed him to explore different musical styles and showcase his versatility as a musician.}} \\
\hline

\texttt{\small{\textbf{\# The Highlights of Taylor Hawkins' Solo Career}}} \\
\texttt{\small{Taylor Hawkins' solo career was marked by a diverse range of musical projects and collaborations, showcasing his versatility as a musician. He released his debut solo album, "Red Light Fever," in 2010, which featured a blend of rock, pop, and power-pop influences. The album's lead single, "Your Shoes," was a testament to Hawkins' love for Cheap Trick-style power-pop and Queen-level arena rock pomp, with harmonized guitar leads, boogie rhythms, and theatrical backing vocals that captured the essence of his rock'n'roll spirit.}} \\
\texttt{\small{Hawkins also contributed to the soundtrack of the 2013 rock film "CBGB," playing the role of Iggy Pop. This role not only highlighted his acting skills but also his ability to embody the energy and charisma of a legendary musician. Additionally, he recorded drums on Vasco Rossi's song "L'uomo più semplice," further expanding his international reach and influence.}} \\
\texttt{\small{Beyond his solo endeavors, Hawkins collaborated with various artists, including Slash, who featured him on the track "Crucify the Dead" from his solo album "Slash." This collaboration demonstrated Hawkins' ability to adapt his drumming style to different musical contexts, adding depth and texture to Slash's hard-hitting rock sound.}} \\
\texttt{\small{Hawkins' solo career also included a period of reflection and personal growth. He candidly documented his struggles with substance abuse in his solo album, which served as a platform for him to share his experiences and emotions with the world. This honesty and authenticity resonated with fans, providing a glimpse into the human side of a musician often seen through the lens of fame and success.}} \\
\texttt{\small{Overall, Taylor Hawkins' solo career was a testament to his musical versatility, creativity, and the depth of his talent. It was a journey that saw him explore different genres, collaborate with esteemed artists, and use his platform to address personal challenges, all while maintaining his status as a beloved member of the Foo Fighters and a respected figure in the music industry.}} \\
\hline

\texttt{\small{\textbf{\# The Cause of Taylor Hawkins' Death}}} \\
\texttt{\small{Taylor Hawkins, the drummer of the band Foo Fighters, passed away on March 25, 2022, at the age of 50. The cause of his death was not immediately revealed, but preliminary toxicology tests conducted by Colombian authorities indicated that he had 10 different substances in his system at the time of his death, including opioids, benzodiazepines, tricyclic antidepressants, and THC. The office of the Attorney General of Colombia stated that they would continue the medical studies to achieve total clarification of the events that led to Hawkins' death, and the investigation is ongoing. As of now, the official cause of death has not been determined, but it is believed that Hawkins' death was likely due to a drug overdose.}} \\
\texttt{\small{Hawkins' friends and colleagues have attempted to make sense of his death in the absence of any real answers. Some have suggested that Hawkins could have had heroin in his system and that a cocaine-like powder was found in his hotel room. However, Hawkins' friend, Yeti, believes resolutely that Hawkins was not using heroin or any other substance because he was supposed to play a concert that night. Three days later, he was dead, and the official cause of death is still unknown.}} \\
\texttt{\small{Meanwhile, a separate report claimed Hawkins' heart "weighed double that of men his age," which also could have contributed to his death. The finger-pointing continued from there, and Rolling Stone later suggested that Hawkins was exhausted by the constant demands of the Foo Fighters.}} \\
\texttt{\small{Hawkins' death sent shockwaves throughout the world, and tributes poured in from fellow musicians, some via social media and others in concert. The Foo Fighters canceled their remaining shows in South America and returned to the United States, where they continued to tour and perform. The band released a statement expressing their grief and love for Hawkins, and they dedicated their shows to him in his memory.}} \\
\texttt{\small{In the aftermath of Hawkins' death, the conversation around drug abuse and mental health has been impacted. Hawkins' struggle with drug addiction and his overdose have sparked discussions about the importance of mental health and the need for better support for musicians and artists who may be struggling with addiction or other mental health issues. The Foo Fighters have also been vocal about the importance of seeking help and getting support when needed, and they have encouraged their fans to do the same.}} \\
\hline

\texttt{\small{\textbf{\# Legacy of Taylor Hawkins}}} \\
\texttt{\small{Taylor Hawkins left a significant legacy in the music industry. His contributions to the band, his passion for music, and his vibrant personality have been remembered by fans, friends, and fellow musicians alike. Hawkins' impact on the Foo Fighters' sound and his ability to connect with audiences through his performances have solidified his place in rock history.}} \\
\texttt{\small{One of Hawkins' most notable achievements was his work on the band's 2021 album "Medicine at Midnight," which won all three awards it was nominated for at the Grammy Awards. The album, featuring nine tracks, showcased Hawkins' versatility and passion, as he navigated both faster, hard-hitting songs like "Cloudspotter" and slower, ballad-like tunes in "Chasing Birds." His dedication to the album and the band's overall success is evident in his words, where he expressed gratitude for the band's loyal fan base and their support during challenging times.}} \\
\texttt{\small{Hawkins' legacy extends beyond his musical contributions. His charismatic presence, athletic performances, and surprising vocal agility were hallmarks of his stagecraft. Friends and family members have shared stories of Hawkins' love for music and his commitment to the Foo Fighters, even as he considered the toll of constant touring. Despite his reservations, Hawkins remained an integral part of the band, his ebullience and unique connection with Dave Grohl electrifying the group's performances.}} \\
\hline

\end{longtable}

\onecolumn
\renewcommand{\arraystretch}{1.5} %

\begin{longtable}{|p{1.0\textwidth}|}
\hline
\texttt{\small{\textbf{Taylor Hawkins}}} \footnotesize{\textit{(Article generated by STORM)}} \\
\hline
\endfirsthead
\hline
\texttt{\small{Oliver Taylor Hawkins (1972-2022) was an American musician, best known as the drummer for the rock band Foo Fighters[1]. Born in Fort Walton, Texas, Hawkins' love for music was ignited at a young age, particularly after watching a Queen concert in 1982[2][3][5]. He kick-started his professional career as the drummer for Alanis Morissette’s band during her world tour for the hit album 'Jagged Little Pill'[8][9]. His talents were recognized by Dave Grohl, who invited him to join the Foo Fighters in 1997, marking the beginning of his impactful tenure with the band[7][8].}} \\
\texttt{\small{Hawkins was celebrated for his versatile drumming style, drawing influence from renowned drummers like Roger Taylor, Neil Peart, Phil Collins, Alex Van Halen, and Stewart Copeland[14]. His performances, marked by a unique energy and aggressive style of play, earned him recognition as one of the top rock drummers of his era[15]. Apart from his role in the Foo Fighters, Hawkins' passion for music saw him involved in numerous side projects and collaborations, cementing his place in the world of rock music[10].}} \\
\texttt{\small{Outside of his professional life, Hawkins was known for his resilience and dedication to his family. Despite personal struggles, including a near-fatal drug overdose in 2001, Hawkins remained committed to his musical career[4][9]. His legacy continues to inspire musicians and fans alike, as his contributions to rock music, coupled with his indomitable spirit, made him an unforgettable icon in the music industry[13].}} \\
\texttt{\small{Hawkins' sudden death in 2022 while on tour in Bogotá, Colombia, sent shockwaves through the music world[34]. Tributes poured in from around the globe, reflecting the respect and admiration Hawkins had garnered during his lifetime[21][31]. His life and career were honored at a star-studded tribute concert in Los Angeles, attesting to the enduring impact of his music[22].}} \\
\hline

\texttt{\small{\textbf{\# Early Life and Background}}} \\
\texttt{\small{Oliver Taylor Hawkins, known as Taylor Hawkins, was born and raised in Fort Walton, Texas[3]. His family moved to Laguna Beach, California when he was four years old[3]. He has two younger siblings, a brother named Jason, and a sister named Heather[3]. As a child, Hawkins was particularly influenced by his paternal grandmother, Josie Hawkins, who had grown up during the Great Depression and lived in Jackson, Mississippi[1].}} \\
\texttt{\small{During his high school days at Laguna Beach High School, from where he graduated in 1990, he became friends with Jon Davison, who later became the lead vocalist of the band Yes[2][3]. His interest in music was nurtured from an early age, particularly after watching a Queen concert in 1982 which inspired him to learn to play the drums[2][5]. He noted that music was a constant presence in his family home[5].}} \\
\texttt{\small{Despite facing certain hardships during his upbringing, including his mother's struggles with "demons", Hawkins pursued his musical ambitions[4]. He credits his older sister Heather for taking care of the family during difficult times[4].}} \\
\texttt{\small{His first major musical experience came from playing drums for Alanis Morissette's album, Jagged Little Pill, and accompanying her on the subsequent tour[3]. This marked the beginning of his professional career in the music industry.}} \\
\hline

\texttt{\small{\textbf{\# Career}}} \\
\texttt{\small{Taylor Hawkins began his professional music career playing in Alanis Morissette’s band during her 18-month world tour in support of the hit album 'Jagged Little Pill' from 1995 to 1997[8][9]. His performances not only in the tour but also in the music videos for “You Oughta Know”, “All I Really Want” and “You Learn” introduced him to the world of rock music and ultimately led to his meeting with Dave Grohl[8]. Throughout this time, Hawkins contributed significantly to the band's sound and performance, transforming the songs from their original drum loop format to a rock-band vibe that resonated with audiences[1][7].}} \\
\texttt{\small{In 1997, Hawkins was asked by Grohl to join the Foo Fighters, an invitation that he readily accepted[7][8]. At the time, Grohl thought it was a long shot to recruit Hawkins given that Morissette was at the height of her career, but Hawkins' desire to be a part of a rock band compelled him to make the move[7]. This marked the beginning of Hawkins' tenure as the drummer of the Foo Fighters, a role that he would play until his passing[6][9].}} \\
\texttt{\small{Apart from his work with Morissette and the Foo Fighters, Hawkins had an array of other musical experiences[10]. He drummed for Sass Jordan before joining Morissette’s touring band[10]. He was part of an ad hoc drum supergroup called SOS Allstars and filled the void for Coheed and Cambria’s 2007 album after their drummer Josh Eppard left the group[10]. In addition, Hawkins formed his own side project, the Coattail Riders, in 2005, through which he recorded his own music and took the project on the road, performing in small clubs despite the Foo Fighters' arena-status[7]. His son, Shane Hawkins, has since taken on his father's legacy, joining the Foo Fighters for a performance during the Boston Calling Music Festival in 2023[6].}} \\
\hline

\texttt{\small{\textbf{\# Musical Style and Influences}}} \\
\texttt{\small{Taylor Hawkins was a profound drummer, with his musical style and influences spreading across a wide array of rock genres[11]. Known for his passionate fandom of groups that came before him, Hawkins regularly expressed his admiration for bands like Rush, Genesis, and the Police, all of which featured some of the greatest drummers in rock history like Neil Peart, Phil Collins, and Stewart Copeland[11].}} \\
\texttt{\small{He was heavily influenced by his love for classic rock, as evidenced by his performances, where he covered songs from bands like Van Halen[11].}} \\
\texttt{\small{Hawkins drew influences from a variety of drumming styles, developing a signature style inspired by greats like Roger Taylor, Neil Peart, Phil Collins, Alex Van Halen, and Stewart Copeland[14]. This distinctive style and influence extended to his drum kit, which incorporated elements like rototoms and concert toms[14].}} \\
\texttt{\small{Beyond his influences, Hawkins had a unique energy that made him stand out as a drummer. His performances were recognized for their power, and he was known for his enthusiastic and aggressive style of play[15]. This earned him recognition as one of the top rock drummers of his time, with his passion for music living on through his performances[14].}} \\
\hline

\texttt{\small{\textbf{\# Personal Life}}} \\
\texttt{\small{Taylor Hawkins married Alison Hawkins, an American celebrity and entrepreneur, in 2005[18]. The couple had three children, Oliver, Annabelle, and Everleigh[19]. Hawkins' commitment to his family was evident; in fact, he even wrote a song for his middle child, Annabelle[9].}} \\
\texttt{\small{In his personal life, Hawkins had also struggled with drug use, which nearly claimed his life in a 2001 overdose[9][7][4]. However, he managed to overcome this challenge, and later expressed gratitude for the experience as a lesson that allowed him to realize the destructive path he was on[7].}} \\
\texttt{\small{Outside of his main role in the Foo Fighters, Hawkins also pursued various side projects including the Birds of Satan, NHC, and Chevy Metal. His motivation for such ventures was a constant drive to create and his love for music[7]. Hawkins was also known for his unabashed fanboy nature, often vocalizing his admiration for fellow musicians and his heroes[7].}} \\
\hline

\end{longtable}

\end{document}